\pdfoutput=1
\documentclass[11pt]{article}

\usepackage{authblk}

\usepackage[]{acl}

\usepackage{times}
\usepackage{latexsym}
\usepackage{booktabs}
\usepackage{hyperref}
\usepackage{graphicx}
\usepackage{multirow}
\usepackage{latexsym}

\usepackage[T1]{fontenc}
\usepackage[utf8]{inputenc}

\usepackage{microtype}

\newcommand{\pg}{\textsc{ParaGen}~}
\newcommand{\pgs}{\textsc{ParaGen}'s~}

\title{\pg: A Parallel Generation Toolkit}

\makeatletter
\def\thanks#1{\protected@xdef\@thanks{\@thanks
        \protect\footnotetext{#1}}}
\makeatother

\author[1]{\textbf{Jiangtao Feng}\textsuperscript{*}\thanks{\textsuperscript{*} Work was done at ByteDance.}}
\author[2]{\textbf{Yi Zhou}}
\author[1]{\textbf{Jun Zhang}\textsuperscript{*}}
\author[2]{\textbf{Xian Qian}}
\author[2]{\textbf{Liwei Wu}}
\author[2]{\textbf{Zhexi Zhang}}
\author[3]{\authorcr \textbf{Yanming Liu}\textsuperscript{*}} 
\author[2]{\textbf{Mingxuan Wang}}
\author[4]{\textbf{Lei Li}\textsuperscript{*}}
\author[5]{\textbf{Hao Zhou}\textsuperscript{*}}

\affil[1]{Shanghai AI Laboratory}
\affil[2]{ByteDance Inc.}
\affil[3]{Shanghai Jiaotong University}
\affil[4]{Univeristy of California Santa Barbara}
\affil[5]{Insititute for AI Industry Research, Tsinghua University}

\begin{document}

\maketitle

\begin{abstract}
\pg is a PyTorch-based NLP toolkit for further development on parallel generation.
\pg provides thirteen types of customizable plugins, helping users to experiment quickly with novel ideas across model architectures, optimization, and learning strategies.
We implement various features, such as unlimited data loading and automatic model selection, to enhance its industrial usage.
ParaGen is now deployed to support various research and industry applications at ByteDance. 
\pg is available at \url{https://github.com/bytedance/ParaGen}.
\end{abstract}

\section{Introduction}
Recently, neural sequence generation model achieve great success~\cite{vaswani2017attention, lewis-etal-2020-bart, liu2020multilingual}.
Among a surge of sequence generation algorithms, parallel generation or non-autoregressive generation methods gain increasing attention on various tasks for high inference speed~\cite{gu2017non, saharia2020non, qian2020glancing, gu2020fully, huang2021non} and competitive performance against auto-regressive transformer~\cite{gu2019levenshtein, chan2020imputer, qian2021volctrans}.
Apart from natural language processing, parallel generation also demonstrates its superiority and scalability on text-to-speech synthesis~\cite{ren2020fastspeech} and high-resolution image synthesis~\cite{chang2022maskgit}.

Several toolkits on sequence generation has been presented for developing sequence generation algorithms, such as FairSeq~\cite{ott2019fairseq}, Tensor2Tensor~\cite{vaswani2018tensor2tensor}, Transformers~\cite{wolf2019huggingface} and OpenNMT~\cite{klein-etal-2017-opennmt}.
These toolkits are mostly born with auto-regressive transformers with maximum likelihood estimation training and are used for research purposes.

\begin{figure}[t]
    \centering
    \includegraphics[width=\linewidth]{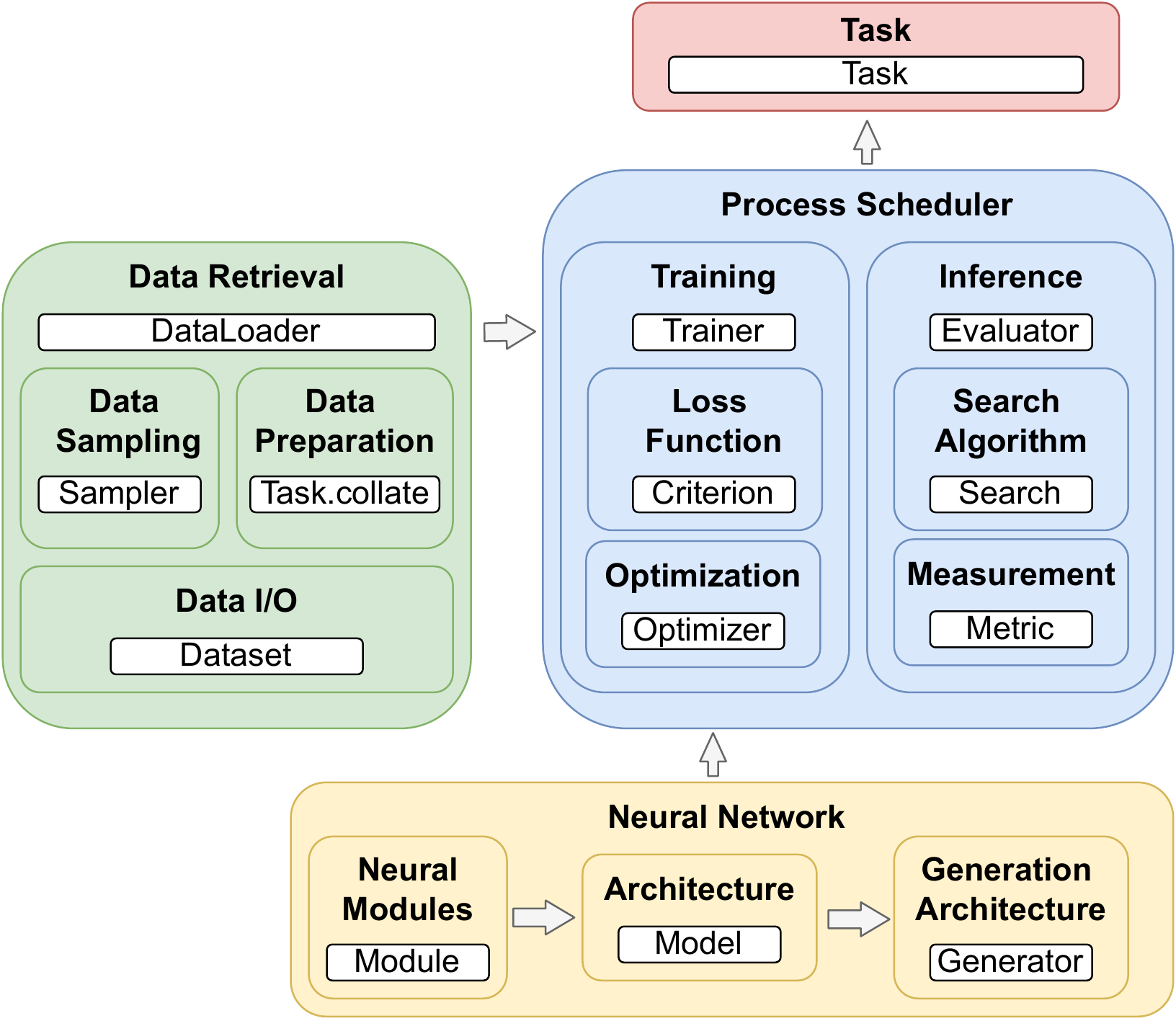}
    \caption{Overall framework of \pg. The data retrieval block is colored with green; the neural network modules are colored with yellow; the process scheduling processing block is colored with bleu; and the task, which dominates all the processes, is colored with red. All the white blocks here are customizable classes in \pg.}
    \label{fig:framework}
\end{figure}

In this paper, we present \pg, an extensible toolkit for parallel generation, which is first developed with Glancing Transformer on WMT-21 Competition~\cite{qian2021volctrans}.
We redesign the code architecture for easy modification on training and decoding methods, such as glancing training~\cite{qian2020glancing}, imitation learning~\cite{wei2019imitation}, inference algorithms~(noisy parallel decoding)~\cite{gu2017non}, and mask-predict decoding~\cite{ghazvininejad2019mask}, which are critical to enhancing parallel generation algorithm developments.
Besides, \pg also suits industrial usage with robust implementations and attractive features, such as unlimited data loading, asynchronized input/output, plug-in Huggingface tokenizers/models~\cite{wolf2019huggingface} and fast training/inference with LightSeq~\cite{wang2020lightseq, wang2021lightseq2}.
Apart from parallel generation, \pg also reproduces typical tasks with step-by-step scripts, such as autoregressive translation~\cite{vaswani2017attention}, text summarization~\cite{lewis-etal-2020-bart}, text classification~\cite{wang2018glue}, and extractive question answering~\cite{rajpurkar2016squad}.
As for large-scale pretraining, \pg supports BERT pretraining~\cite{devlin2018bert}, and multilingual translation with mBART pretraining~\cite{liu2020multilingual}.
\pg is now deployed to support various research and industrial applications at ByteDance.

\section{Architecture Design}
The overall architecture of \pg is shown as Figure~\ref{fig:framework}.
\pg consists of four main functional blocks: data, model, trainer, and evaluator.
The data block focuses on data input, processing, sampling, and loading; the model block consists of neural models in training and inference; the trainer is implemented for scheduling the training process; the evaluator defines the evaluation metrics.
Compared with the previous frameworks, we offer 13 types of plug-ins across the three blocks, which makes \pg more extensible for experimenting with new ideas.

\subsection{Data}

We design the data organization block on four base concepts, including reading, preprocessing, sampling strategy and loading, deriving four customizable class or functions respectively, i.e. \texttt{Dataset}, \texttt{Data Processing}, \texttt{Sampler} and \texttt{DataLoader}.
We address \pg's data process paradigm along with two key topics: online-offline data processing and unlimited data loading challenge.

\paragraph{Dataset} 
The \texttt{Dataset} instances read data and organize it to a \texttt{dict}-format object, despite their storage format on disks.
Users are allowed to develop their on \texttt{Dataset} class for customization usage by implementing \texttt{load} and \texttt{callback} functions.
Currently, \pg supports data stored in various formats, including raw texts, parallel texts, and JSON files.
The \texttt{Dataset}s as well as other classes in \pg co-work with an underlying \texttt{io} module to suit different file systems, reading and writing data on a local disk or a Hadoop file system.
It is worth noting that the \texttt{io} module is also modularized and extensible to suit data input/output under more scenarios.
Besides, we also develop \texttt{StreamingDataset}, reading data in a streaming way.
The \texttt{StreamingDataset} can read extremely large-scale data with constant memory consumption, making it extensible to industrial usage.

\paragraph{Data Processing} 
Data preprocessing, such Byte-Pair Encoding~\cite{sennrich-etal-2016-neural}, is critical to sequence generation and varies from task to task. 
To enhance task-specific data preprocessing, \pg provides interfaces within \texttt{Task} class to allow customization. 
The data processing is roughly divided into two categories, offline data processing as \texttt{data\_collate\_fn} and online data processing \texttt{collate\_fn}.
The \texttt{data\_collate\_fn} refers to offline data processing and proceeds before the training/inference stage start with input from \texttt{Dataset}.
Thus data processed by \texttt{data\_collate\_fn} remains unchanged during the training/inference process, which speeds up training and inference by eliminating repeated data processing.
The \texttt{collate\_fn} is designed as online processing to enhance flexibility and to allow users to adjust data processing strategies, such as batching, during training and inference.
% We use a flat way for data processing for its easy implementation.
% Moreover, we also include a post-processing function to process data 
We believe the combination of offline and online data processing would make data processing more flexible and extensible.

\paragraph{Sampler} 
The sampling strategy is a non-negligible algorithm in the online data processing.
Although PyTorch provides a base class of sampling strategy, it is still often ignored by existing generation frameworks.
\pg allows users to develop their sampling strategies by implementing a \texttt{Sampler} instance to decide how data are organized into batches.
% We here offer a base sampler class to allow customization of data sampling strategy.
A technical challenge of incorporating customizable sampling strategies is their compatibility with the feature of unlimited data loading.
We solve this problem in the \texttt{DataLoader} with a cache mechanism.

\paragraph{DataLoader} 
\texttt{DataLoader} is the final stage of data processing and the beginning of neural model processing, acting as a bridge to connect data and neural models.
It can also be viewed as a coordinator of data processing.
It first fetches a batch of samples, according to the sampling strategy determined by \texttt{Sampler}, from data memory with offline processed data.
Then it sends the data batches to online data processing, which becomes a private object of \texttt{DataLoader} instance at initialization, and gets a batch to feed the neural network.
However, in the original PyTorch, \texttt{DataLoader} is incompatible with streaming data loading.
We extend the dataloader and implement a \texttt{StreamingDataLoader} to read data streamingly, further featuring unlimited data loading.

\subsection{Module}

A key principle of designing \pg's model block is based on the concept separation of training and inference.
Unlike classification models, sequence generation ones become different when they are applied in training and inference.
For example, a sequence generation model is trained in a teacher-forcing way whilst it generates sequences with the help of the beam search algorithm.
Thus we defines four sub-modules for \pg model block: \texttt{Model}, \texttt{Generator}, \texttt{Criterion} and \texttt{Search}.
\texttt{Model} and \texttt{Criterion} are often used in training process whereas \texttt{Generator} and \texttt{Search} are in inference.
% It would be helpful for further neural model performance optimization in training and inference.
% Because under some scenarios, the computational graph of a neural model is different during two stages.
We argue that the training-inference separation of neural models would benefit the downstream neural model optimization~\cite{wang2020lightseq} in industrial usage without harming flexibility in developing new models.

\paragraph{Model}
We implement the architecture of neural models with learnable parameters by inheriting \pgs \texttt{Model} class.
Similar to the models in the existing, it consumes a batch of samples and produces logits over the predicted target.
We provide several popular implementations, including parallel generation, autoregressive sequence generation, extraction model, and sequence classification/regression.
Besides, Huggingface models at present are widely-used for their large-scale pre-training models, and we implemented a Huggingface model wrapper to make it compatible in \pg.

\paragraph{Generator}
In \pg, we advocate \texttt{Generator}, instead of the original \texttt{Model}, to apply to the inference stage.
\texttt{Generator} is designed as a wrapper to \texttt{Model} with extra decoding algorithms, such as beam search and greedy search in autoregressive sequence generation~\cite{vaswani2017attention}, noisy padding mask in parallel generation~\cite{gu2017non}, and even extraction algorithms in extractive tasks.
Although the extra decoding algorithms could also be implemented as post-processing, they would benefit from tensor computation on GPUs, which further speeds up the computations.
It is also recommended in \pg to have decoding algorithms on GPUs and post-processing on CPUs work with each other to achieve the tradeoff between flexibility and speedup.
We here argue that it is essential for industrial usage to separate \texttt{Generator} from \texttt{Model}, because \texttt{Generator}, which is gradient-free, requires more elaborated and extreme optimization to enhance efficiency.
Previous study~\cite{wang2020lightseq} shows that joint optimization on neural models and decoding algorithms achieves significant speedup and reduces GPU memory consumption.
Nevertheless, such joint optimization does not suit \texttt{Model} in training.
In \pg, we implement \texttt{Generators} for sequence extraction, auto regressive sequence generation and parallel sequence generation.

\paragraph{Criterion} 
Like previous frameworks, we define \texttt{Criterion} class as the objective functions across various tasks.
It measures the divergence between predicted logits and golden reference.
From the \texttt{Criterion}, we compute the gradients of all the learnable parameters for optimization.
We also allow neural models, such as Huggingface models, to compute loss by themselves.
Due to the modularization in \pg, \texttt{Criterion}s can be combined to enhance multi-task learning.

\paragraph{Search} 
Search or decoding algorithms, such as beam search and noise padding mask, are critical to sequence generation. 
We modularize \texttt{Search} to support users in developing their awesome sequence search algorithms, for both autoregressive or parallel generation, more than existing ones.
The \texttt{Search} algorithms act as a part of \texttt{Generator}, co-working with \texttt{Model} to produce final sequences.

\subsection{Trainer}

One big difference between \pg and existing learning framework is customizing the training process.
Recent research shows that a neural model with the same architecture trained with a well-designed training strategy performs significantly better.
Customizing a \texttt{Trainer} helps the users to experiment with training strategies conveniently. 
The \texttt{Trainer} formulates the whole training process and includes several types of customization: loss computation, optimizer, and rate scheduler. 

\paragraph{Loss computation} 
\texttt{Trainer} leaves an interface \texttt{forward\_loss} for implementation of loss computation.
Elaboration on loss computation is critical to deep learning algorithms to enhance models' performance.
For examples, 
a) GLAT~\cite{qian2020glancing} can be used for computing a three-stage objective, glancing at the target, modifying neural network inputs/targets, and learning;
b) FreeLB~\cite{zhu2019freelb} adopts an adversarial gradient to inject to neural model to learn it robustly;
c) CoNT~\cite{an2022cont} leverage a generation process to adopt contrastive learning to enhance sequence generation.
Thus we believe that customization on loss computation frees developers from the stereotyped training process and encourage new experimental training algorithms.

\paragraph{Optimizer} 
\pg provides \texttt{Optimizer} customization following the original PyTorch.
All the optimizers implemented in PyTorch and Huggingface can be used directly, and experimental optimizers are also encouraged.
The coordination of \texttt{Optimizer} to advanced optimization algorithms, such as mix-precision training, apex support, and distributed training, is automatic. 

\paragraph{Rate Scheduler}
We implement a functional tool called \texttt{Rate Scheduler} to define how a rate typed as an integer or float is scheduled.
The rate in \pg could be any hyper-parameter beyond the learning rate, allowing users to schedule their training process more flexibly.
In design, the original PyTorch treats the optimizer as an attribute to the learning rate scheduler, but \pg does this differently by setting the scheduled learning rate as a part of the optimizer.
The learning rate is actively scheduled through the interaction between \texttt{Optimizer} and \texttt{Trainer}.

\subsection{Evaluator}

\texttt{Evaluator} formulates the overall evaluation process in \pg, supporting customization on two aspects, including data sets and metrics.
In evaluations, we compute the Cartesian product of data sets and metrics to obtain several performance scores. The scores are averaged to obtain an overall judgment on the current neural model.
These scores will further return to \texttt{Trainer} for model selection.

\paragraph{Metric}
We provide \texttt{Metric} for independent evaluation of the divergence between predicted hypotheses and ground-truth references.
\texttt{Metric} in \pg can be designed as lexical metrics~(BLEU~\cite{papineni-etal-2002-bleu} and Rouge~\cite{lin2004rouge}), numeric metrics~(Accuracy and F1-score), and model-based metrics~(BERT-score~\cite{zhang2019bertscore})

\subsection{Difference to FairSeq}

A close sequence generation toolkit to \pg is FairSeq~\cite{ott2019fairseq}.
\pg differs from FairSeq in four aspects: 
a) FairSeq is for general purpose sequence generation, while \pg is carefully designed for \textit{parallel generation};
% b) FairSeq supports 5 types of customizable modules whereas \pg offers 13 classes;
% c) FairSeq is process-oriented by filling codes into a training or inference process while \pg is designed task-oriented by picking up modules and organizing them to complete scripts;
% d) FairSeq is mainly used for research, but 
b) Fairseq's standard pipeline requires binarizing the data first, which is extremely fast and efficient for training. Such a way may make on-the-fly data manipulation a bit difficult. \pg allows reading the raw data and modifying it dynamically in the training loop, without sacrificing the speed.
c) FairSeq supports 5 types of customizable modules whereas \pg offers 13 classes. Fairseq provides a unified training loop, \pg disentangles the training and inference process into independent modules~(\texttt{Trainer}, \texttt{Evaluator}), so that the customization for each task is more user-friendly.
d) \pg implements more features specially designed for industrial usage.

\section{Implementation \& Features}

\pg implements fundamental functions, such as distributed training on multiple machines and GPUs (with Horovod~\cite{sergeev2018horovod}), mix-precision training (with apex\footnote{https://github.com/NVIDIA/apex}), incremental decoding for the autoregressive models, breakpoint resuming, out-of-memory recovery, early stopping, and accumulated gradients.
Moreover, we also implement advanced functions, including unlimited data loading, automatic model selection, multi-task training, and fast training/inference with LightSeq.
In this section, we focus on these advanced features.

\paragraph{Unlimited Data Loading}
Pretrained models has show its strong capability in generalizing to new tasks~\cite{devlin2018bert,liu2019roberta,radford2019language,lewis-etal-2020-bart,2020t5,floridi2020gpt}.
However, such pre-trained models usually demand billions, trillions, and even more of the training data.
Moreover, for industrial usage, the large amount of internal data is a challenge for data loading.
In \pg, we implement a \texttt{StreamingDataLoader} that reads data from disk in a streaming way with the file input stream.
It also features data distribution with multi-GPU training and local shuffling for data batching.
With the help of \texttt{StreamingDataLoader}, \pg can read unlimited data with limited memory usage.

\paragraph{Automatic Model Selection}
\pg automatically selects the best models with a customized assessment metric during the training process.
By providing the trainer with an \texttt{assess\_by} arguments as \texttt{\{DATA\_NAME\}.\{METRIC\_NAME\}}, the trainer picks up the checkpoints that performs the best on \texttt{\{DATA\_NAME\}} with respect to \texttt{\{METRIC\_NAME\}}.
Note that once a checkpoint is selected to save, \pg also saves the average of $k$ checkpoints before and marks the averaged checkpoints as \texttt{best\_avg}\footnote{We find it performs better compared with the average of the last-$k$ checkpoints after the training process ends.}.

\paragraph{Asynchronized Input/Output} 
\pg uses an asynchronized input/output implementation for reading and writing to maximize the utility of GPU resources.
Asychronized output is important for model selection.
Because the computation of \texttt{best\_avg} checkpoints costs a long time to finish, especially for large models in industrial applications.

\paragraph{Multi-Task Learning}
\pg provides an easy-to-use way for multi-task learning.
\pg implements a \texttt{MultiTaskCriterion} which automatically combines a list of criteria.

\paragraph{Plug-in Hugginface Tokenizers and Models}
Huggingface~\cite{wolf2019huggingface} is a widely used pre-trained model library and demonstrates its effectiveness among various tasks.
In \pg, Hugginface can be directly used in an import-register way.
It allows researchers to develop algorithms upon pre-trained models and to use advanced features provided in \pg.

\paragraph{Fast Training and Inference with LightSeq}
In \pg, researchers and developers can use LigthSeq easier to speedup their transformer models/modules in training~\cite{wang2021lightseq2} and inference~\cite{wang2020lightseq}.
Without understanding the details of LightSeq, users can speed up their transformer models/modules by simply appending a \texttt{LS} prefix to the model/module class name in the configuration.

\section{Reproducibility}
\pg can be used for various tasks beyond parallel generation.
We provide reproducible results and scripts on six benchmarks on \pg, including: 
a) glancing transformer on WMT14 En-De;
b) transformer on IWSLT14 De-En, WMT14 En-De, and WMT14 En-Fr;
b) transformer on Multi-News and XSum;
c) mBART on Multilingual Translation;
d) plug-in Huggingface on SQuAD 1.1;
e) BERT pretraining and fine-tuning on GLUE benchmark.
For simplicity, we eliminate the details of reproduction configurations and hyper-parameters.
The reproduction results are shown in Appendix~\ref{sec:reproduce} and scripts can be found in our repository.

\section{Conclusions}

We present the ParaGen toolkit for parallel sequence generation.
It supports 13 types of modules for customization and advocates a plug-in usage for further development.
Its robust implementation and features enhance the research algorithm design and industrial development.
In the future, we will create more plugins to extend \pg to more research areas.

\bibliography{custom}
\bibliographystyle{acl_natbib}

\newpage
\appendix
\section{Reproducible Results}
\label{sec:reproduce}

We implement typical models on various tasks:
a) glancing transformer on WMT14 En-De;
b) transformer on IWSLT14 De-En, WMT14 En-De, and WMT14 En-Fr;
b) transformer on Multi-News and XSum;
c) mBART on Multilingual Translation;
d) plug-in Huggingface on SQuAD 1.1;
e) BERT pretraining and fine-tuning on GLUE benchmark.
These models are implemented with reproducible scripts.
The detailed reproducible results are shown in the rest of the section.

\subsection{Glancing Transformer on WMT14 En-De}

We trained Glancing Transformer on Transformer-AT distilled data.
We report sacrebleu~\cite{post-2018-call} and tokenized BLEU~\cite{papineni2002bleu} for completeness.

\begin{table}[htbp]
    \centering
    \footnotesize
    \begin{tabular}{c|c|c|c}
        \toprule
        Model & sacrebleu & tok bleu \\
        \midrule
        GLAT & 24.40 & 24.98 \\
        GLAT + avg-ckpt & 24.58 & 25.29 \\
        \bottomrule
    \end{tabular}
    \caption{BLEUs on distilled WMT14 En-De. GLAT+avg-ckpt is the average checkpoint among the last 10 checkpoints.}
\end{table}

\begin{table}[htbp]
    \centering
    \begin{tabular}{c|c|c|c}
        \toprule
        Task & Model & sacrebleu & tok bleu \\
        \midrule
        IWSLT14 De-En & small & 33.1 & 34.5 \\
        WMT14 En-De & base & 26.9 & 27.5 \\
        WMT14 En-De & big & 27.7 & 28.4 \\
        WMT14 En-Fr & big & 40.3 & 43.3 \\
        \bottomrule
    \end{tabular}
    \caption{BLEUs on machine translation benchmarks.}
\end{table}

\subsection{Transformer on Machine Translation}

We implement widely-used transformer architecture and provide its results on IWSLT/WMT benchmarks.
For IWSLT14 De-En, we use transformer-small architecture; for WMT14 En-De and WMT14 En-Fr, we use transformer-big architecture.
Similar to Glancing Transformer, we report its results with sacrebleu and tokenized bleu.

\subsection{mBART on Multilingual Translation}

\pg provides implementation to pre-train mBART from scratch.
We finetune the pre-trained mBART on XX-en translation tasks.

\begin{table}[htbp]
    \centering
    \begin{tabular}{c|c|c}
        \toprule
        Language Pairs & mBART & mBART + avg-ckpt \\
        \midrule
        de-en & 41.45 & 41.84 \\
        fr-en & 39.09 & 39.41 \\
        ja-en & 21.68 & 22.84 \\
        pl-en & 32.03 & 32.50 \\
        ro-en & 36.75 & 37.24 \\
        mn-en & 12.75 & 14.14 \\
        hi-en & 26.34 & 27.78 \\
        \bottomrule
    \end{tabular}
    \caption{Results on XX-en translation with mBART pre-trained from scratch.}
    \label{tab:my_label}
\end{table}

\subsection{Abstractive Text Summarization}

\begin{table}[htbp]
    \small
    \centering
    \begin{tabular}{c|c|c|c|c}
        \toprule
        task & model & rouge-1 & rouge-2 & rouge-n \\
        \midrule
        \multirow{2}{*}{Multi-News} & tr-base & 33.59 & 5.91 & 30.71 \\
        & BART-base & 46.80 & 17.93 & 43.01 \\
        XSum & bart-base & 42.49 & 19.52 & 34.37 \\
        \bottomrule
    \end{tabular}
    \caption{Results on abstractive text summarization.}
    \label{tab:my_label}
\end{table}

\subsection{Question Answering}

\pg also provides results on extractive question answering with plug-in Huggingface models.
We test various pre-trained models in Huggingface on SQuAD 1.1.

\begin{table}[!htbp]
    \centering
    \begin{tabular}{c|c|c}
        \toprule
        Model & F1 & Exact Match \\
        \midrule
        BERT-base-uncased & 88.31 & 81.02 \\
        BERT-base-cased & 88.34 & 81.11 \\
        BERT-large-cased & 90.73 & 83.87 \\
        RoBERTa-base & 91.88 & 85.34 \\
        RoBERTa-large & 93.44 & 87.26 \\ 
        BART-base & 91.27 & 84.52 \\
        BART-large & 93.14 & 86.70 \\
        \bottomrule
    \end{tabular}
    \caption{Plug-in Huggingface Pretrained Models on SQuAD 1.1}
    \label{tab:my_label}
\end{table}

\subsection{BERT}

\pg supports training BERT from scratch with customized data.
We train BERT on two data sets, news and wibo, and test it on GLUE benchmarks~\cite{wang2018glue}.
We use standard evaluation metrics for each task.

\begin{table*}[!htbp]
    \centering
    \begin{tabular}{l|c|c|c}
        \toprule
        Task & Metric & news & wibo \\
        \midrule
        CoLA & Matthews' Corr & 61.82 & 58.36 \\
        SST-2 & Accuracy & 93.00 & 91.86 \\
        STS-B & Pearson/Spearman Corr & 86.97/86.88 & 87.95/87.65 \\
        MRPC & F1/Accuracy & 91.13/87.75 & 91.13/87.50 \\
        QQP & F1/Accuracy & 87.95/91.05 & 88.05/91.16 \\
        MNLI-m/mm & Accuracy & 83.42/83.28 & 83.24/83.06 \\
        QNLI & Accuracy & 89.22 & 90.81 \\
        RTE & Accuracy & 67.15 & 61.37 \\
        \bottomrule
    \end{tabular}
    \caption{Results on GLUE by BERT pre-trained on news and wibo dataset.}
\end{table*}

\end{document}